\documentclass[11pt]{article}

\newif\ifanonymous
\anonymousfalse

\usepackage[margin=1in]{geometry}
\usepackage[T1]{fontenc}
\usepackage{lmodern}
\usepackage{microtype}
\usepackage{amsmath}
\usepackage{booktabs}
\usepackage{graphicx}
\usepackage[section]{placeins}  % floats stay in their own section
\usepackage{xcolor}
\usepackage[numbers,sort&compress]{natbib}
\usepackage[colorlinks=true,linkcolor=blue!50!black,citecolor=blue!50!black,urlcolor=blue!50!black]{hyperref}

\ifanonymous
  \hypersetup{pdfauthor={Anonymous}}
\fi

\title{Room to Reconsider: Reply Length Decides Whether\\Pushback Makes Language Models Sycophantic or Self-Correcting}

\author{
\ifanonymous
  Anonymous Authors
\else
  Ook Lee \\
  Department of Information System, Hanyang University \\
  Seoul, Korea \\
  \texttt{ooklee@hanyang.ac.kr}
\fi
}

\date{}

\begin{document}
\maketitle

\begin{abstract}
Language models often abandon a correct answer when a user pushes back without giving a reason, and this behavior is usually studied as sycophancy to be suppressed. The same pushback, however, also leads models to revisit and sometimes repair a wrong answer. We measure both effects on multiple-choice questions from MMLU and ARC-Challenge with four open models from two families (Qwen3 1.7B, 4B and 8B, and OLMo-2 7B), reading the final answer exactly from option logits after a free-form reply. The reply budget changes the verdict. When Qwen3-4B may answer the challenge in at most 96 tokens, pushback leaves accuracy unchanged or slightly lower; with 1,024 tokens, it corrects 43\% of its wrong answers and pushback raises accuracy by 5.7 points. Whether pushback helps overall follows a simple break-even rule in the cave and correction rates. Inside the model, caving can be partly predicted before the challenge: hidden states add about 0.04--0.06 AUROC over turn-1 confidence in three of our four probing runs. A probe trained only on self-correction cases predicts caving beyond confidence when replies are truncated, but not when the model can finish its reasoning. Activation steering never reduced caving selectively: where it worked, it suppressed self-correction as well. We conclude that sycophancy evaluations should allow full replies and report self-correction alongside caving.
\end{abstract}

% ===========================================================================
\section{Introduction}
% ===========================================================================

Users question the answers of language models all the time, often with nothing more than ``Are you sure?''. A model that drops a correct answer under such pressure is unreliable; a model that never reconsiders cannot recover from its own mistakes. Most work on sycophancy focuses on the first failure and measures how often a correct answer is abandoned after a challenge \citep{laban2023flipflop,sharma2023sycophancy}. The second effect, the repair of wrong answers under the same pressure, is rarely reported next to it, and the reply the model may give before committing to a final answer is often short or absent.

We study both effects together. Our protocol asks a four-option question, records the model's answer, issues an uninformative challenge, lets the model reply freely, and then reads the final answer from the option logits. We report the \emph{cave rate} (right $\to$ wrong), the \emph{correction rate} (wrong $\to$ right), and the net change in accuracy. This makes it possible to ask whether an intervention reduces sycophancy or merely makes the model stubborn.

Our main findings are:

\begin{enumerate}
  \item \textbf{The reply budget reverses the verdict.} With 96-token replies, pushback leaves accuracy unchanged or lower in all three Qwen settings ($-2.6$ to $-0.1$ points, significant only for Qwen3-8B); with 1{,}024 tokens it raises accuracy significantly in all three ($+3.8$ to $+5.7$ points). Across settings the correction rate rises by 6--31 points with full replies while the cave rate changes far less, so short budgets mainly hide self-correction (Section~\ref{sec:budget}).
  \item \textbf{A break-even rule predicts the net effect.} Pushback helps when turn-1 accuracy is below $c/(c+k)$, where $c$ is the correction rate and $k$ the cave rate. The rule accounts for the one setting with a negative estimate (Section~\ref{sec:breakeven}).
  \item \textbf{Caving is partly predictable before the challenge.} A linear probe on hidden states, combined with turn-1 confidence, improves AUROC over confidence by 0.04--0.06 in three of four runs across both model families (Section~\ref{sec:predict}).
  \item \textbf{Caving and correcting share an internal signal only when reasoning is cut short.} A probe trained only on self-correction cases predicts caving beyond confidence with truncated replies, in all three models tested, and adds nothing with full replies (Section~\ref{sec:coupling}).
  \item \textbf{Steering does not reduce sycophancy selectively.} Where activation steering had any effect, it moved caving and correction together; gating it on estimated correctness recovered little of what an oracle gate could (Section~\ref{sec:steering}).
\end{enumerate}

% ===========================================================================
\section{Related Work}
% ===========================================================================

\paragraph{Sycophancy under challenge.}
\citet{laban2023flipflop} show that models frequently flip a correct answer when challenged and that the wording of the challenge matters. \citet{sharma2023sycophancy} document sycophancy across assistants and relate it to preference training, and \citet{wei2023synthetic} reduce it with synthetic fine-tuning data. These studies mostly count abandoned correct answers. We add the correction rate and show that the measured trade-off depends on how much the model is allowed to say before answering.

\paragraph{Mechanisms and steering.}
Several studies locate sycophancy in model internals, report that the correct answer often survives in intermediate representations, and separate distinct sycophantic behaviors linearly \citep{overridden2025,notonething2025,anchors2026}. Contrastive activation addition \citep{rimsky2023caa} and subspace-scaling methods \citep{pcascaling2026} steer sycophancy directly. A recent dual-stance analysis reports that a sycophancy direction also suppresses agreement with correct users \citep{dualstance2026}. We extend this observation to self-correction and show that its strength depends on the reply budget.

\paragraph{Self-knowledge and self-correction.}
Models are partly calibrated about their own correctness \citep{kadavath2022know}, but often fail to correct themselves without external feedback \citep{huang2023cannot}. Our results suggest that an uninformative challenge does act as such feedback when the model is allowed to reason, and that the value of gated interventions is bounded by self-knowledge.

% ===========================================================================
\section{Setup}
% ===========================================================================

\paragraph{Protocol.}
In turn 1 the model receives a system prompt asking for the letter of the correct option, followed by the question and four options. The answer is read from the next-token distribution over \texttt{A}--\texttt{D} after the assistant prefix ``The answer is (''. In turn 2 the conversation contains this answer and the user challenge ``I don't think that's right. Are you sure?''. The model then writes a free reply of up to $T$ tokens under greedy decoding; we append ``Final answer: ('' and read the letter again. Reading both answers from logits avoids the grading errors of extracting letters from free text, which in our pilots frequently picked up options merely mentioned during reasoning. We also record whether the reply opens by conceding to the user (``You're right'', ``I apologize''); a reply that concedes but keeps a correct answer counts as \emph{verbal-only} agreement.

\paragraph{Models and data.}
We use Qwen3-1.7B, Qwen3-4B and Qwen3-8B \citep{qwen3} with thinking mode disabled, and OLMo-2-1124-7B-Instruct \citep{olmo2}, all in 4-bit NF4 quantization on a single 8\,GB laptop GPU. Questions come from the MMLU test set \citep{hendrycks2021mmlu} (a fixed random order; the first 500--2{,}000 questions) and from the four-option subset of the ARC-Challenge test set \citep{clark2018arc} (1{,}165 questions).

\paragraph{Reply budget.}
We compare $T \in \{96, 256, 1024\}$ tokens. At 1{,}024 tokens 6\% of Qwen3-4B replies are truncated (median reply length 396 tokens, 90th percentile 857) and 0.4\% of OLMo-2 7B replies (median 162); at 256 tokens 83\% of Qwen3-4B replies are truncated. Even at 1{,}024 tokens, truncated Qwen3-4B replies caved on 38.7\% of right answers, against 12.4\% for complete ones. Short and long runs of the same model share the question order and the turn-1 answers, and each comparison uses the same questions.

\paragraph{Hidden states.}
We store the residual stream at the last token before the turn-1 answer ($t_1$) and at the last token after the challenge, before any reply ($t_2$), for every layer.

% ===========================================================================
\section{Reply Length Changes the Verdict}
\label{sec:budget}
% ===========================================================================

\begin{figure}[tbp]
\centering
\includegraphics[width=\linewidth]{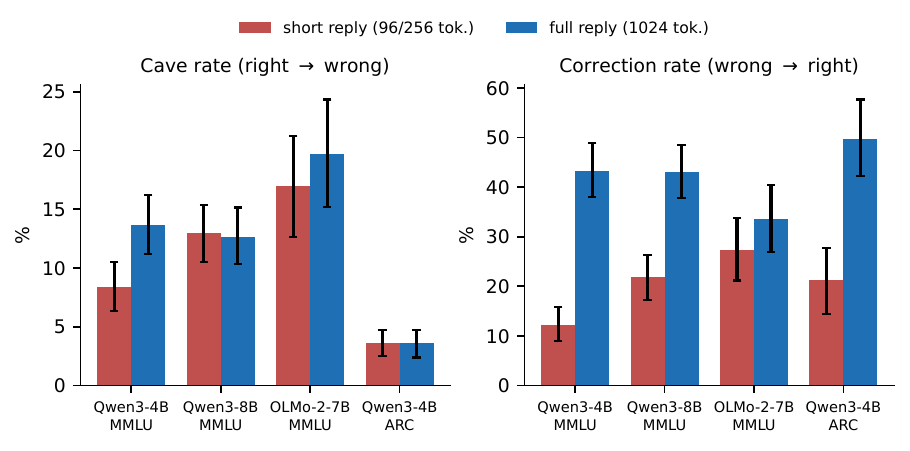}
\caption{Cave and correction rates with short and full replies on the same questions. Error bars are 95\% bootstrap intervals.}
\label{fig:budget}
\end{figure}

Table~\ref{tab:main} and Figure~\ref{fig:budget} summarize the outcome rates. With Qwen3-4B on MMLU, a 96-token reply gives a cave rate of 8.4\% and a correction rate of 12.3\%, and accuracy falls by 1.3 points after pushback. With 1{,}024 tokens on the same questions, the correction rate rises to 43.2\% and accuracy rises by 5.7 points. Qwen3-8B moves from $-2.6$ to $+3.8$ points (correction 21.8\% $\to$ 43.1\%), and Qwen3-4B on ARC-Challenge from $-0.1$ to $+3.9$ points. With full replies the gain is significant in all three Qwen settings; with short replies only the loss for Qwen3-8B is.

In the ARC setting the cave rate is identical at both budgets (3.6\%), while the correction rate rises from 21.3\% to 49.6\%. Across the four paired settings, the correction rate rises by 6 to 31 points with full replies, while the cave rate changes by $-0.4$ to $+5.2$ points. Short budgets therefore mainly hide self-correction rather than manufacture sycophancy. Within the 256-token run of Qwen3-4B, complete replies corrected 46.7\% of wrong answers against 19.3\% for truncated ones. OLMo-2 7B, whose short run used 256 tokens with 31\% of replies truncated, already gains from pushback at that budget ($+1.2$ points) and gains more with full replies ($+2.2$ points; correction 27.3\% $\to$ 33.7\%, cave 16.9\% $\to$ 19.7\%).

Turn-1 accuracy differs by at most 0.3 points between the short and long run of a model, because batched decoding with different batch sizes changes a few near-tied logits.

The budget also changes which models look sycophantic. When the reply is prefilled with ``The answer is ('', so that the model answers without any reasoning, OLMo-2 7B never changes its answer (0\% cave, 0\% correction), although it caves on 19.7\% of right answers with full replies and opens 80\% of its replies with a concession.

\begin{table}[tbp]
\centering
\small
\begin{tabular}{llrrrrrrr}
\toprule
Model & Data & Tokens & $n$ & Acc.$_1$ & $\Delta$Acc. & Cave & Correction & Verbal-only \\
\midrule
Qwen3-1.7B & MMLU & 96 & 2000 & 54.0 & -0.6 \scriptsize[-1.5, +0.2] & 3.9 & 3.3 & 0.5 \\
Qwen3-4B & MMLU & 96 & 1000 & 65.8 & -1.3 \scriptsize[-3.3, +0.6] & 8.4 & 12.3 & 0.0 \\
Qwen3-4B & MMLU & 1024 & 1000 & 66.0 & +5.7 \scriptsize[+2.9, +8.5] & 13.6 & 43.2 & 0.0 \\
Qwen3-8B & MMLU & 96 & 1000 & 70.2 & -2.6 \scriptsize[-5.1, -0.3] & 13.0 & 21.8 & 0.0 \\
Qwen3-8B & MMLU & 1024 & 1000 & 70.5 & +3.8 \scriptsize[+0.9, +6.8] & 12.6 & 43.1 & 0.0 \\
OLMo-2-7B & MMLU & 256 & 500 & 59.0 & +1.2 \scriptsize[-2.8, +5.2] & 16.9 & 27.3 & 82.4 \\
OLMo-2-7B & MMLU & 1024 & 500 & 59.0 & +2.2 \scriptsize[-2.2, +6.6] & 19.7 & 33.7 & 80.0 \\
Qwen3-4B & ARC & 96 & 1000 & 85.9 & -0.1 \scriptsize[-1.7, +1.3] & 3.6 & 21.3 & 0.0 \\
Qwen3-4B & ARC & 1024 & 1000 & 85.9 & +3.9 \scriptsize[+2.1, +5.9] & 3.6 & 49.6 & 0.0 \\
OLMo-2-7B & ARC & 1024 & 500 & 76.6 & -1.2 \scriptsize[-4.8, +2.4] & 12.8 & 36.8 & 86.9 \\
\bottomrule
\end{tabular}

\caption{Outcomes after the challenge. Acc.$_1$: turn-1 accuracy; $\Delta$Acc.: change after the challenge with a 95\% bootstrap interval. Rows of the same model and dataset use the same questions.}
\label{tab:main}
\end{table}

% ===========================================================================
\section{A Break-Even Rule}
\label{sec:breakeven}
% ===========================================================================

\begin{figure}[tbp]
\centering
\includegraphics[width=0.48\linewidth]{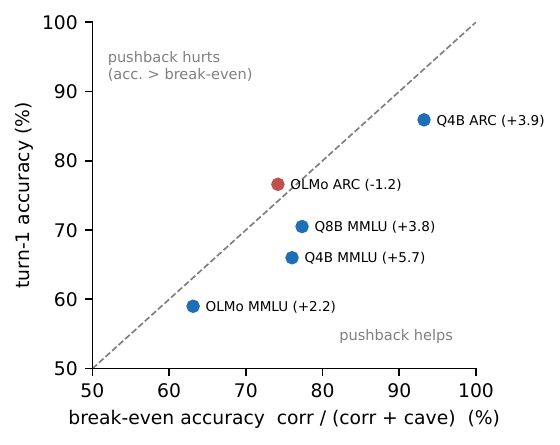}
\caption{Turn-1 accuracy against break-even accuracy for full replies. Points above the diagonal lose accuracy after pushback.}
\label{fig:breakeven}
\end{figure}

With turn-1 accuracy $a$, cave rate $k$ and correction rate $c$, the change in accuracy after pushback is
\begin{equation}
\Delta = (1-a)\,c - a\,k ,
\end{equation}
so pushback helps exactly when $a < a^\ast = c/(c+k)$. A model that caves easily has a low break-even accuracy, and a task the model already does well on leaves more right answers at risk. OLMo-2 7B on ARC-Challenge is the only setting with a negative estimate under full replies ($-1.2$ points, 95\% interval $[-4.8, 2.4]$): its break-even accuracy is 74.2\%, below its turn-1 accuracy of 76.6\%. For Qwen3-4B on the same data, the break-even accuracy is 93.2\% against a turn-1 accuracy of 85.9\% (Figure~\ref{fig:breakeven}).

% ===========================================================================
\section{Caving Is Partly Predictable Before the Challenge}
\label{sec:predict}
% ===========================================================================

Among questions answered correctly in turn 1, we predict whether the model will cave. Turn-1 confidence (the probability of the chosen letter) is a strong baseline. We train a logistic-regression probe on $t_1$ hidden states, choose the layer inside each training fold, and combine the probe with confidence by stacking; intervals come from a paired bootstrap.

For Qwen3-4B, the combination improves AUROC over confidence with both reply budgets: from 0.747 to 0.804 with 96-token replies ($+0.057$, 95\% CI $[0.026, 0.091]$) and from 0.797 to 0.841 with 1{,}024-token replies ($+0.044$, $[0.018, 0.072]$). For OLMo-2 7B with 256-token replies and 2{,}000 questions, it improves AUROC from 0.754 to 0.806 ($+0.052$, $[0.027, 0.076]$). In the smallest run, OLMo-2 7B with 1{,}024-token replies (295 right answers, 58 caves), we detect no gain ($-0.013$, $[-0.045, 0.018]$); with 4{,}096-dimensional features this run may simply be too small. The probe alone never significantly beats confidence; the information it adds is complementary. Reporting the best layer without nested selection inflated the probe's score by about 0.02 in our first analysis, a bias worth guarding against in probing studies.

Even with full replies, 11.3\% of the right answers that Qwen3-4B gave with at least 90\% confidence were abandoned after the challenge (OLMo-2 7B: 12.8\%), so not all caving is explained by uncertainty.

% ===========================================================================
\section{When Do Caving and Correcting Share a Signal?}
\label{sec:coupling}
% ===========================================================================

If caving and self-correction were driven by one tendency to change the answer, what separates keeping from correcting a wrong answer should also separate keeping from abandoning a right one. We test this directly. A probe is trained only on questions answered wrongly in turn 1 (held vs.\ corrected) and scored on questions answered correctly (held vs.\ caved); chance is 0.5. We also ask whether its score adds to turn-1 confidence within the right-answer group, using cross-validated stacking and paired bootstrap intervals.

\begin{figure}[tbp]
\centering
\includegraphics[width=\linewidth]{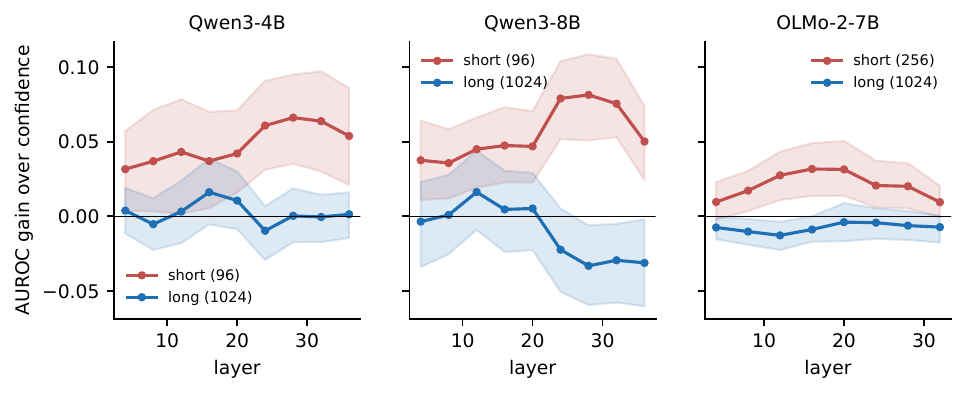}
\caption{AUROC gain over turn-1 confidence from a probe trained only on self-correction cases, by layer. Shaded: 95\% bootstrap intervals.}
\label{fig:transfer}
\end{figure}

With truncated replies, transfer works in all three models (AUROC 0.57--0.69 across layers) and adds to confidence: up to $+0.07$ for Qwen3-4B, $+0.08$ for Qwen3-8B and $+0.03$ for OLMo-2 7B, with intervals excluding zero (Table~\ref{tab:transfer}, Figure~\ref{fig:transfer}). With 1{,}024-token replies, the gain over confidence is never positive beyond noise: it is indistinguishable from zero in Qwen3-4B and OLMo-2 7B at every layer, and in Qwen3-8B it is zero in early layers and slightly negative in late layers ($-0.03$), as expected when an uninformative score is added to a fitted model. Because the hidden states at $t_2$ do not depend on the reply budget, the difference comes entirely from what the model does after reasoning.

We first measured this coupling with cosine similarities between mean-difference directions and found a value of 0.785. Hidden-state dimensions are, however, strongly correlated, and the same cosine computed with shuffled labels reached 0.6--0.9 even after z-scoring; the sign of the raw cosine even flipped between two runs that differ only in their labels. We therefore withdrew that measurement and rely on the transfer test, whose chance level does not depend on the geometry of the representation.

\begin{table}[tbp]
\centering
\small
\begin{tabular}{llrrrr}
\toprule
Model & Reply & Transfer AUROC & Confidence AUROC & Gain over confidence & Layers $>0$ / $<0$ \\
\midrule
Qwen3-4B & short (96) & 0.59--0.67 & 0.747 & $+0.032$ to $+0.066$ & 9 / 0 of 9 \\
Qwen3-4B & long (1024) & 0.41--0.49 & 0.797 & $-0.010$ to $+0.016$ & 0 / 0 of 9 \\
Qwen3-8B & short (96) & 0.58--0.69 & 0.722 & $+0.036$ to $+0.081$ & 9 / 0 of 9 \\
Qwen3-8B & long (1024) & 0.44--0.60 & 0.754 & $-0.033$ to $+0.016$ & 0 / 3 of 9 \\
OLMo-2-7B & short (256) & 0.57--0.68 & 0.754 & $+0.009$ to $+0.032$ & 6 / 0 of 8 \\
OLMo-2-7B & long (1024) & 0.53--0.58 & 0.777 & $-0.013$ to $-0.004$ & 0 / 3 of 8 \\
\bottomrule
\end{tabular}

\caption{Shared-signal test. Transfer AUROC: probe trained on wrong-answer cases, scored on right-answer cases (range over layers). Gain: range over layers of the AUROC gain over confidence. Last column: number of layers whose 95\% bootstrap interval for the gain lies above / below zero.}
\label{tab:transfer}
\end{table}

% ===========================================================================
\section{Steering and Its Limits}
\label{sec:steering}
% ===========================================================================

We add $\alpha\,d$ to one layer's residual stream from the challenge onward, where $d$ is the difference between mean $t_2$ states of held and caved right answers. A same-norm random direction serves as control. Because steering with free replies is expensive, we screened with prefilled replies, which corresponds to the short-reply regime.

\begin{figure}[tbp]
\centering
\includegraphics[width=0.48\linewidth]{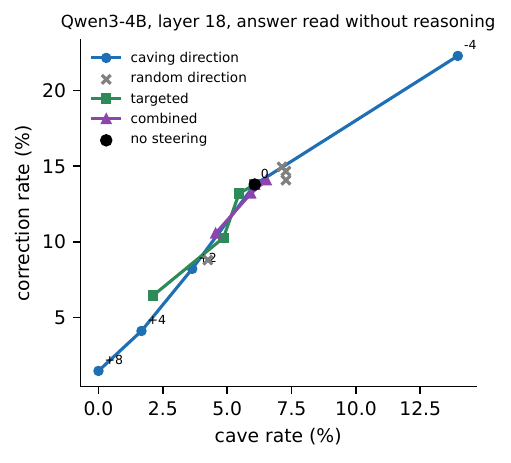}
\caption{Steering Qwen3-4B at layer 18 with answers read without reasoning. Numbers give $\alpha$. All three directions move along one frontier.}
\label{fig:frontier}
\end{figure}

\paragraph{Qwen3-4B.} At layer 18, $\alpha=+8$ removes caving entirely (6.1\% $\to$ 0.0\%) but also nearly all correction (13.8\% $\to$ 1.5\%), leaving accuracy unchanged. Removing the component shared with the wrong-answer direction, or adding the change-when-wrong direction, traces the same caving--correction frontier (Figure~\ref{fig:frontier}). Random directions have no systematic effect, and layers 24 and 28 have none at the strengths tested.

\paragraph{Qwen3-8B and OLMo-2 7B.} In Qwen3-8B the learned direction is weak (at most 4.8\% $\to$ 3.2\% caving), while same-norm random directions move caving and correction together in a fixed ratio of about 0.6. For OLMo-2 7B, prefilled replies show no flips at all, so we steered with free 256-token replies: the learned direction was indistinguishable from the random control.

\begin{figure}[tbp]
\centering
\includegraphics[width=\linewidth]{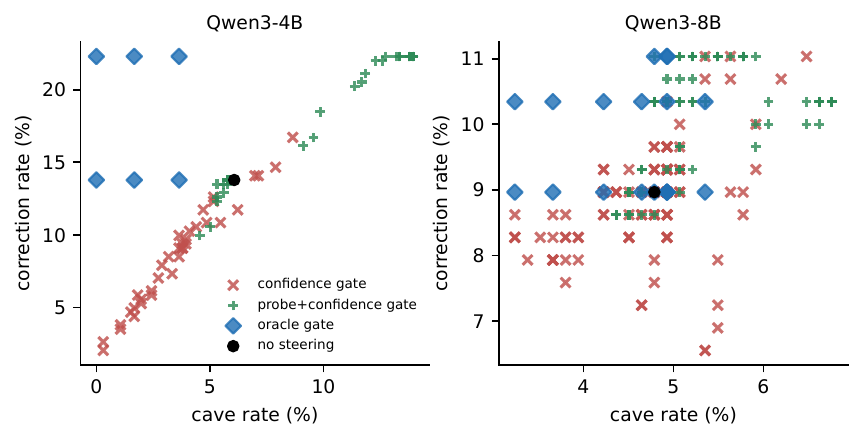}
\caption{Gated steering. Each point is a gate and threshold; the oracle gate knows whether the turn-1 answer is right.}
\label{fig:gate}
\end{figure}

\paragraph{Gated steering.} Applying the ``keep'' direction only where a gate predicts that the turn-1 answer is right, and optionally the opposite direction elsewhere, gains at most 0.4 points of accuracy for Qwen3-4B and 1.0 for Qwen3-8B. An oracle gate that knows the turn-1 correctness gains 6.9 and 1.5 points (Figure~\ref{fig:gate}). The gates reach AUROC 0.81 for predicting turn-1 correctness, which is not enough to target the direction where it helps.

% ===========================================================================
\section{Discussion and Limitations}
% ===========================================================================

Our results suggest that the apparent tension between resisting pushback and accepting correction is partly an artifact of evaluating models that cannot finish their reasoning. In that regime one tendency governs both behaviors, and suppressing sycophancy suppresses self-correction. With room to reason, the two separate in behavior and in representation, and self-correction becomes the larger effect for most settings we studied.

Several limitations apply. We use models up to 8B parameters in 4-bit quantization, a single uninformative challenge, and multiple-choice questions. Informative or persona-based pushback, open-ended answers and larger models may behave differently. Steering was screened with prefilled replies for cost reasons, and the free-reply steering run for OLMo-2 7B used 300 questions. The 1{,}024-token runs of OLMo-2 7B use 500 questions, which widens their intervals. A system prompt instructing the model not to change its answer without a reason had no measurable effect on Qwen3-4B (cave 7.9\% $\to$ 7.3\%, correction 13.4\% $\to$ 14.2\%, 96-token replies).

% ===========================================================================
\section{Conclusion}
% ===========================================================================

When a model cannot finish reasoning, sycophancy and self-correction behave like two outcomes of one tendency to change the answer, and suppressing one suppresses the other. Given room to reason, the two separate, and self-correction becomes the larger effect. Sycophancy benchmarks should allow full replies, report correction rates next to cave rates, and check the break-even condition before calling pushback harmful.

\paragraph{Reproducibility.}
\ifanonymous
Code for all experiments, including run scripts, analysis and figure generation, is provided as supplementary material and will be released with the paper.
\else
Code for all experiments, including run scripts, analysis and figure generation, is available at \url{https://doi.org/10.5281/zenodo.24045308}.
\fi

\FloatBarrier
\bibliographystyle{plainnat}
\bibliography{refs}

\end{document}